\pdfoutput=1
\documentclass[11pt,a4paper]{article}
\usepackage[hyperref]{acl2020}
\usepackage{times}
\usepackage{latexsym}

\usepackage{microtype}

\aclfinalcopy % Uncomment this line for the final submission
\usepackage{amsmath}
\usepackage{multirow}
\usepackage{CJKutf8}
\usepackage{xcolor}
\usepackage{mathtools}
\usepackage{subfig}

\title{Zero-Shot Paraphrase Generation with Multilingual Language Models}

\author{
	Yinpeng Guo\textsuperscript{\rm 1}, Yi Liao\textsuperscript{\rm 1}, Xin Jiang\textsuperscript{\rm 1}, Qing Zhang\textsuperscript{\rm 2}, Yibo Zhang\textsuperscript{\rm 2}, Qun Liu\textsuperscript{\rm 1} \\
	\textsuperscript{\rm 1}Huawei Noah's Ark Lab \\
	\textsuperscript{\rm 2}Intelligence Engineering Department, Huawei Consumer Business Group \\
	\texttt{\{guo.yinpeng, liao.yi, jiang.xin, zhangqing49,} \\
    \texttt{yibo.cheung, qun.liu\}@huawei.com}\\
	}

\date{}

\begin{document}
\begin{CJK*}{UTF8}{gbsn}
	
\maketitle

\begin{abstract}
Leveraging multilingual parallel texts to automatically generate paraphrases has drawn much attention as size of high-quality paraphrase corpus is limited. Round-trip translation, also known as the \textit{pivoting} method, is a typical approach to this end. However, we notice that the pivoting process involves multiple machine translation models and is likely to incur semantic drift during the two-step translations. In this paper, inspired by the Transformer-based language models, we propose a simple and unified paraphrasing model, which is purely trained on multilingual parallel data and can conduct \textit{zero-shot} paraphrase generation in one step. Compared with the pivoting approach, paraphrases generated by our model is more semantically similar to the input sentence. Moreover, since our model shares the same architecture as GPT~\cite{Radford2018}, we are able to pre-train the model on large-scale unparallel corpus, which further improves the fluency of the output sentences. In addition, we introduce the mechanism of denoising auto-encoder (DAE) to improve diversity and robustness of the model. Experimental results show that our model surpasses the pivoting method in terms of relevance, diversity, fluency and efficiency.

\end{abstract}

\section{Introduction}
% paraphrase generation
Paraphrasing is to express the same meaning using different expressions. Paraphrase generation plays an important role in various natural language processing (NLP) tasks such as response diversification in dialogue system, query reformulation in information retrieval, and data augmentation in machine translation. Recently, models based on Seq2Seq learning~\citep{Sutskever2014} have achieved the state-of-the-art results on paraphrase generation. Most of these models~\cite{prakash-etal-2016-neural,Cao2017,Gupta2018,Li2018,Li2019} focus on training the paraphrasing models based on a paraphrase corpus, which contains a number of pairs of paraphrases. However, high-quality paraphrases are usually difficult to acquire in practice, which becomes the major limitation of these methods. Therefore, we focus on \textit{zero-shot paraphrase generation} approach in this paper, which aims to generate paraphrases without requiring a paraphrase corpus.

\begin{figure}[t]
	\centering
	\includegraphics[width=.7\columnwidth]{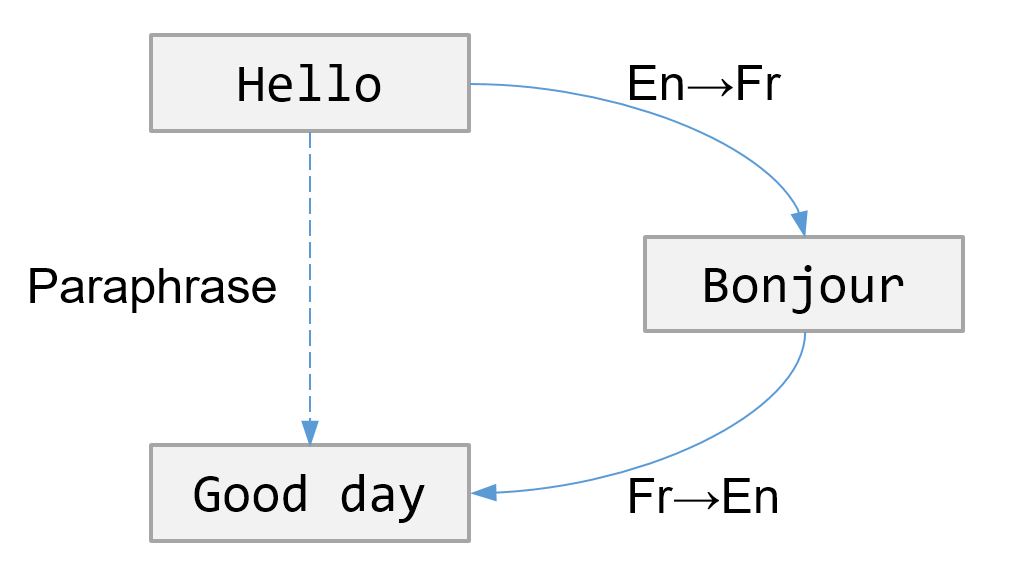} % Reduce the figure size so that it is slightly narrower than the column. Don't use precise values for figure width.This setup will avoid overfull boxes.
	\caption{Paraphrase generation via round-trip translation.}
	\label{fig:round-trip}
\end{figure}

A natural choice is to leverage the bilingual or multilingual parallel data used in machine translation, which are of great quantity and quality. The basic assumption is that if two sentences in one language (e.g., English) have the same translation in another language (e.g., French), they are assumed to have the same meaning, i.e., they are paraphrases of each other. Therefore, one typical solution for paraphrasing in one language is to \textit{pivot} over a translation in another language. Specifically, it is implemented as the \textit{round-trip} translation, where the input sentence is translated into a foreign sentence, then back-translated into a sentence in the same language as input~\cite{Mallinson2018}. The process is shown in Figure~\ref{fig:round-trip}. Apparently, two machine translation systems (English$\rightarrow$French and French$\leftarrow$English) are needed to conduct the generation of a paraphrase.

Although the pivoting approach works in general, there are several intrinsic defects. First, the round-trip system can hardly explore all the paths of paraphrasing, since it is pivoted through the finite intermedia outputs of a translation system. More formally, let $Z$ denote the meaning representation of a sentence $X$, and finding paraphrases of $X$ can be treated as sampling another sentence $Y$ conditioning on the representation $Z$. Ideally, paraphrases should be generated by following $P(Y|X) = \int_{Z} P(Y|Z)P(Z|X)dZ$, which is marginalized over all possible values of $Z$. However, in the round-trip translation, only one or several $Z$s are sampled from the machine translation system $P(Z|X)$, which can lead to an inaccurate approximation of the whole distribution and is prone to the problem of \textit{semantic drift} due to the sampling variances. Second, the results are determined by the pre-existing translation systems, and it is difficult to optimize the pipeline end-to-end. Last, the system is not efficient especially at the inference stage, because it needs two rounds of translation decoding.

%In order to analyze these defects formally, let $S^*$ denote meaning or semantic representation of a sentence $X$. Finding the paraphrases of $X$ can be treated as sampling specific sentences $Y$ from a distribution conditioned on $S^*$.
%Thus we formulate the key problem in paraphrase task as finding the optimal $S^*$ of $P(S|X)$.
%
%For pivoting method, an ideal solution would be optimizing $P(S|X) = \int P(S|Z)P(Z|X)dZ$, where $Z$ is a latent sentence in the pivoting language.
%However, instead of modeling the integral over all the feasible latent $Z$ drawn from $P(Z|X)$, typical round-trip models just pick the optimal value of $P(Z|X)$ by translating the source sentence into a sentence in foreign language in the first step. Such approximation results in the inaccurate modeling of $P(S|X)$, which we view as \textit{drifting of semantics}.

To address these issues, we propose a single-step zero-shot paraphrase generation model, which can be trained on machine translation corpora in an end-to-end fashion. Unlike the pivoting approach, our proposed model does not involve explicit translation between multiple languages.
Instead, it directly learns the paraphrasing distribution $P(Y|X)$ from the parallel data sampled from $P(Z|X)$ and $P(Y|Z)$. Specifically, we build a Transformer-based~\cite{Vaswani2017} language model, which is trained on the concatenated bilingual parallel sentences with language indicators. At inference stage, given a input sentence in a particular language, the model is guided to generate sentences in the same language, which are deemed as paraphrases of the input. Our model is simple and compact, and can empirically reduce the risk of semantic drift to a large extent. Moreover, we can initialize our model with generative pre-training (GPT)~\cite{Radford2018} on monolingual data, which can benefit the generation in low-resource languages. Finally, we borrow the idea of denoising auto-encoder (DAE) to further enhance robustness in paraphrase generation.

% results & contribution
We conduct experiments on zero-shot paraphrase generation task, and find that the proposed model significantly outperforms the pivoting approach in terms of both automatic and human evaluations. Meanwhile, the training and inference cost are largely reduced compared to the pivot-based methods which involves multiple systems.

%The major contributions of this work include:

%1) A simple and unified approach for zero-shot paraphrase generation is proposed, and it naturally works on multiple languages;

%2) The proposed model outperforms previous pivoting method in terms of relevance, diversity, fluency and efficiency.

\section{Methodology}

\begin{figure*}[t]
    \centering
    \subfloat[Multilingual Language Model Training.]{\includegraphics[width=0.47\textwidth]{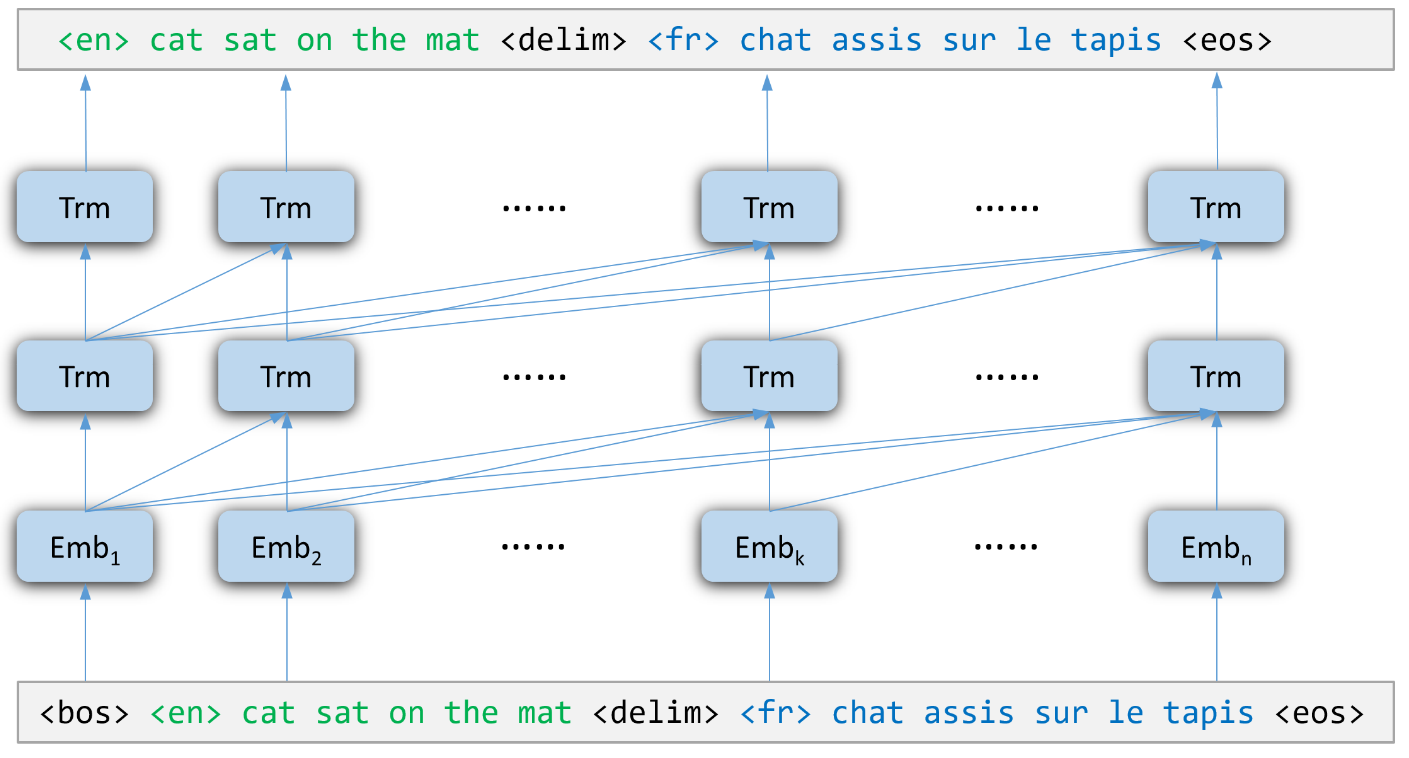}\label{fig:tlm-train}}
    \hfill
    \subfloat[Zero-Shot Paraphrase Generation.]{\includegraphics[width=0.47\textwidth]{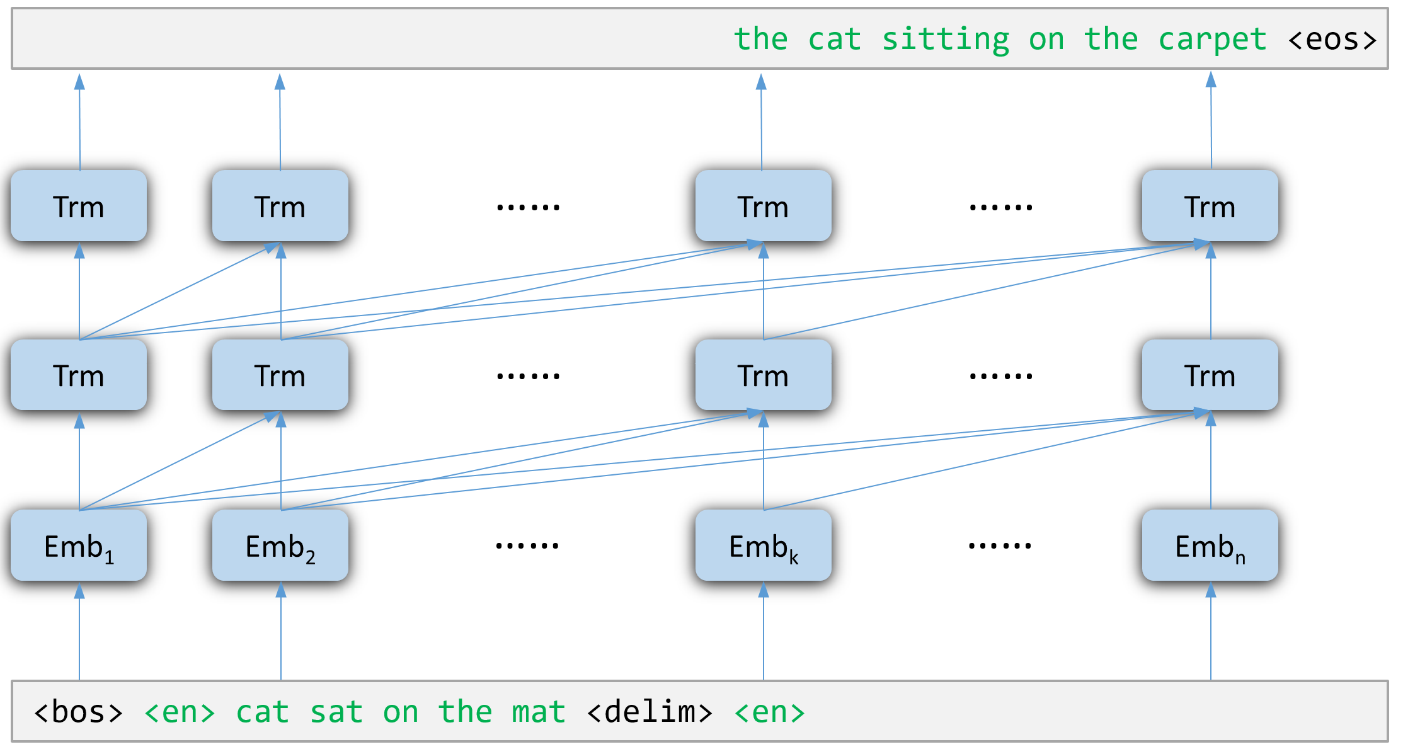}\label{fig:tlm-decode}}
    \caption{Paraphrase generation via multilingual language model training.}\label{fig:tlm}
\end{figure*}

\subsection{Transformer-based Language Model}
Transformer-based language model (TLM) is a neural language model constructed with a stack of Transformer \textit{decoder} layers~\cite{Vaswani2017}. Given a sequence of tokens, TLM is trained with maximizing the likelihood:
\begin{equation} \label{eq-tlm}
L(X)=\sum_{i=1}^n\log{P(x_i|x_{1,\ldots,{i-1}};\theta)}
\end{equation}
where \(X=[x_1,x_2,\ldots,x_n]\) is a sentence in a language (e.g., English), and \(\theta\) denotes the parameters of the model. Each Transformer layer is composed of multi-head self-attention, layer normalization and a feed-forward network. We refer reader to the original paper for details of each component. Formally, the decoding probability is given by
\begin{align}\label{eq-transformer}
&[e_1, \ldots, e_{i-1}] = [W_e x_1 + p_1,\ldots, W_e x_{i-1}+p_{i-1}],\nonumber\\
&[h_1, \ldots, h_{i-1}] = \mathrm{Transformer}([e_1, \ldots, e_{i-1}]), \nonumber\\
&P(x_i|x_{1,\ldots,{i-1}};\theta) = \mathrm{Softmax}(W_o h_{i-1}),
\end{align}
where $x_i$ denotes the token embedding, $p_i$ denote the positional embedding and $h_i$ denotes the output states of the $i$-th token, and $W_e$ and $W_o$ are the input and output embedding matrices.

Although TLM is normally employed to model monolingual sequences, there is no barrier to utilize TLM to model sequences in multiple languages. In this paper, inspired by~\citet{lample2019cross}, we concatenate pairs of sentences from bilingual parallel corpora (e.g., English$\rightarrow$French) as training instances to the model. Let $X$ and $Y$ denote the parallel sentences in two different languages, the training objective becomes
\begin{align} \label{eq-bitlm}
L(X,&Y)=\sum_{i=1}^n\log{P(x_i|x_{1,\ldots,{i-1}};\theta)}  \nonumber \\
 &+ \sum_{j=1}^{m}\log{P(y_j|x_{1,\ldots,{n}},y_{1,\ldots,j-1};\theta)}.
\end{align}
% This bilingual language model can be regarded as a simplified sequence-to-sequence (Seq2Seq) model without explicit encoder and decoder modules. One advantage of such formulation is that it naturally integrate with language models pre-trained on unparalleled corpus.
This bilingual language model can be regarded as the decoder-only model compared to the traditional encoder-decoder model. It has been proved to work effectively on monolingual text-to-text generation tasks such as summarization \cite{Peter2018}. The advantages of such architecture include less model parameters, easier optimization and potential better performance for longer sequences. Furthermore, it naturally integrates with language models pre-training on monolingual corpus.

For each input sequence of concatenated sentences, we add special tokens \textit{$\langle$bos$\rangle$} and \textit{$\langle$eos$\rangle$} at the beginning and the end, and \textit{$\langle$delim$\rangle$} in between the sentences. Moreover, at the beginning of each sentence, we add a special token as its language identifier, for instance, \textit{$\langle$en$\rangle$} for English, \textit{$\langle$fr$\rangle$} for French. One example of English$\rightarrow$French training sequence is ``\textit{$\langle$bos$\rangle$ $\langle$en$\rangle$ cat sat on the mat $\langle$delim$\rangle$ $\langle$fr$\rangle$ chat assis sur le tapis $\langle$eos$\rangle$}".

At inference stage, the model predicts the next word as the conventional auto-regressive model:
\begin{equation} \label{eq-decode}
\hat{y}_j \sim P(y_j|x_{1,...,n},y_{1,...,j-1};\theta)
\end{equation}

%\begin{figure}[t]
%	\centering
%	\includegraphics[width=1\columnwidth]{figure1} % Reduce the figure size so that it is slightly narrower than the column. Don't use precise values for figure width.This setup will avoid overfull boxes.
%	\caption{Zero-shot paraphrase generation based on parallel translation corpus.}
%	\label{fig1}
%\end{figure}

\subsection{Zero-shot Paraphrase Generation}

We train the bilingual language model on multiple bilingual corpora, for example, English$\leftrightarrow$French and German$\leftrightarrow$Chinese. Once the language model has been trained, we can conduct zero-shot paraphrase generation based on the model. Specifically, given an input sentence that is fed into the language model, we set the output language identifier the same as input, and then simply conduct decoding to generate paraphrases of the input sentence.

Figure~\ref{fig:tlm} illustrates the training and decoding process of our model. In the training stage, the model is trained to sequentially generate the input sentence and its translation in a specific language. Training is conducted in the way of teacher-forcing. In the decoding stage, after an English sentence ``\textit{$\langle$bos$\rangle$ $\langle$en$\rangle$ cat sat on the mat $\langle$delim$\rangle$}" is fed to the model, we intentionally set the output language identifier as ``\textit{$\langle$en$\rangle$}", in order to guide the model to continue to generate English words. At the same time, since the model has been trained on translation corpus, it implicitly learns to keep the semantic meaning of the output sentence the same as the input. Accordingly, the model will probably generate the paraphrases of the input sentence, such as ``\textit{the cat sitting on the carpet $\langle$eos$\rangle$}".

It should be noted our model can obviously be trained on parallel paraphrase data without any modification. But in this paper, we will mainly focus on the research and evaluation in the zero-shot learning setting.

In the preliminary experiments of zero-shot paraphrasing, we find the model does not perform consistently well and sometimes fails to generate the words in the correct language as indicated by the language identifier. Similar phenomenon has been observed in the research of zero-shot neural machine translation~\cite{sestorain2018zero,arivazhagan2019missing,Gu2019}, which is referred as the \textit{degeneracy} problem by~\citet{Gu2019}. To address these problems in zero-shot paraphrase generation, we propose several techniques to improve the quality and diversity of the model as follows.

\subsubsection{Language Embeddings}~\label{sec:lan}
The language identifier prior to the sentence does not always guarantee the language of the sequences generated by the model. In order to keep the language consistency, we introduce language embeddings, where each language is assigned a specific vector representation. Supposing that the language embedding for the $i$-th token in a sentence is $a_i$, we concatenate the language embedding with the Transformer output states and feed it to the softmax layer for predicting each token:
\begin{equation} \label{eq-decode}
P(y_j|x_{1,...,n},y_{1,...,j-1};\theta) = \mathrm{Softmax}(W_o [h_{j}, a_j])
\end{equation}
We empirically demonstrate that the language embedding added to each tokens can effectively guide the model to generate sentences in the required language. Note that we still let the model to learn the output distribution for each language rather than simply restricting the vocabularies of output space. This offers flexibility to handle coding switching cases commonly seen in real-world data, e.g., English words could also appear in French sentences.

\subsubsection{Pre-Training on Monolingual Corpora}
Language model pre-training has shown its effectiveness in language generation tasks such as machine translation, text summarization and generative question answering~\cite{radford2019language,dong2019unified,song2019mass}. It is particularly helpful to the low/zero-resource tasks since the knowledge learned from large-scale monolingual corpus can be transferred to downstream tasks via the pre-training-then-fine-tuning approach. Since our model for paraphrase generation shares the same architecture as the language model, we are able to pre-train the model on massive monolingual data.

Pre-training on monolingual data is conducted in the same way as training on parallel data, except that each training example contains only one sentence with the beginning/end of sequence tokens and the language identifier. The language embeddings are also employed. The pre-training objective is the same as Equation (\ref{eq-tlm}).

In our experiments, we first pre-train the model on monolingual corpora of multiple languages respectively, and then fine-tune the model on parallel corpora.

\subsubsection{Denoising Auto-Encoder}
We adopt the idea of denoising auto-encoder (DAE) to further improve the robustness of our paraphrasing model. DAE is originally proposed to learn intermediate representations that are robust to partial corruption of the inputs in training auto-encoders~\cite{Vincent2008}. Specifically, the initial input \(X\) is first partially corrupted as \(\tilde{X}\), which can be treated as sampling from a noise distribution \(\tilde{X}\sim{q(\tilde{X}|X)}\). Then, an auto-encoder is trained to recover the original $X$ from the noisy input \(\tilde{X}\) by minimizing the reconstruction error. In the applications of text generation~\cite{freitag2018unsupervised} and machine translation~\cite{Kim2018}, DAE has shown to be able to learn representations that are more robust to input noises and also generalize to unseen examples.

Inspired by \cite{Kim2018}, we directly inject three different types of noises into input sentence that are commonly encountered in real applications.

\noindent 1) \textit{Deletion}: We randomly delete 1\% tokens from source sentences, for example, ``\textit{cat sat on the mat $\mapsto$ cat on the mat.}"

\noindent 2) \textit{Insertion}: We insert a random token into source sentences in 1\% random positions, for example, ``\textit{cat sat on the mat $\mapsto$ cat sat on red the mat.}"

\noindent 3) \textit{Reordering}: We randomly swap 1\% tokens in source sentences, and keep the distance between tokens being swapped within 5. ``\textit{cat sat on the mat $\mapsto$ mat sat on the cat.}"

By introducing such noises into the input sentences while keeping the target sentences clean in training, our model can be more stable in generating paraphrases and generalisable to unseen sentences in the training corpus. The training objective with DAE becomes
\begin{align} \label{eq-bitlm-dae}
L(X,&Y)= L(X) + L(Y|\tilde{X})q(\tilde{X}|X) \nonumber \\
 &=\sum_{i=1}^n\log{P(x_i|x_{1,\ldots,{i-1}};\theta)} \nonumber \\
 &+ \sum_{j=1}^{m}\log{P(y_j|\tilde{x}_{1,\ldots,{n}},y_{1,\ldots,j-1};\theta)}.
\end{align}

Once the model is trained, we generate paraphrases of a given sentence based on $P(Y|X;\theta)$.

\section{Experiments}

\begin{table*}[ht]
	\caption{Statistics of training data (\#sentences).}
	\centering
	%\resizebox{2\columnwidth}{!}{
		\begin{tabular}{lcccccc}
			\hline
			 & En$\leftrightarrow$Es & En$\leftrightarrow$Ru & En$\leftrightarrow$Zh & Es$\leftrightarrow$Ru & Es$\leftrightarrow$Zh  & Ru$\leftrightarrow$Zh\\
			\hline
			%			\textbf{OpenSubtitles-18} & 61.4M(*0.19) & 25.9M(*0.45) & 11.2M & 18.8M(*0.56) & 8.5M & 4.4M(*2.18) \\
			OpenSubtitles & 11.7M & 11.7M & 11.2M & 10.5M & 8.5M & 9.6M \\
			
			MultiUN & 11.4M & 11.7M & 9.6M & 10.6M & 9.8M & 9.6M \\
			\hline
			%			\textbf{Total} & 72.8M & 37.6M & 20.8M & 29.4M & 18.3M & 14M \\
			Total & 23.1M & 23.4M & 20.8M & 21.1M & 18.3M & 19.2M \\
			\hline
		\end{tabular}
	%}
	\label{table1}
\end{table*}

\begin{figure*}[!htb]
	\centering
	\includegraphics[width=2\columnwidth]{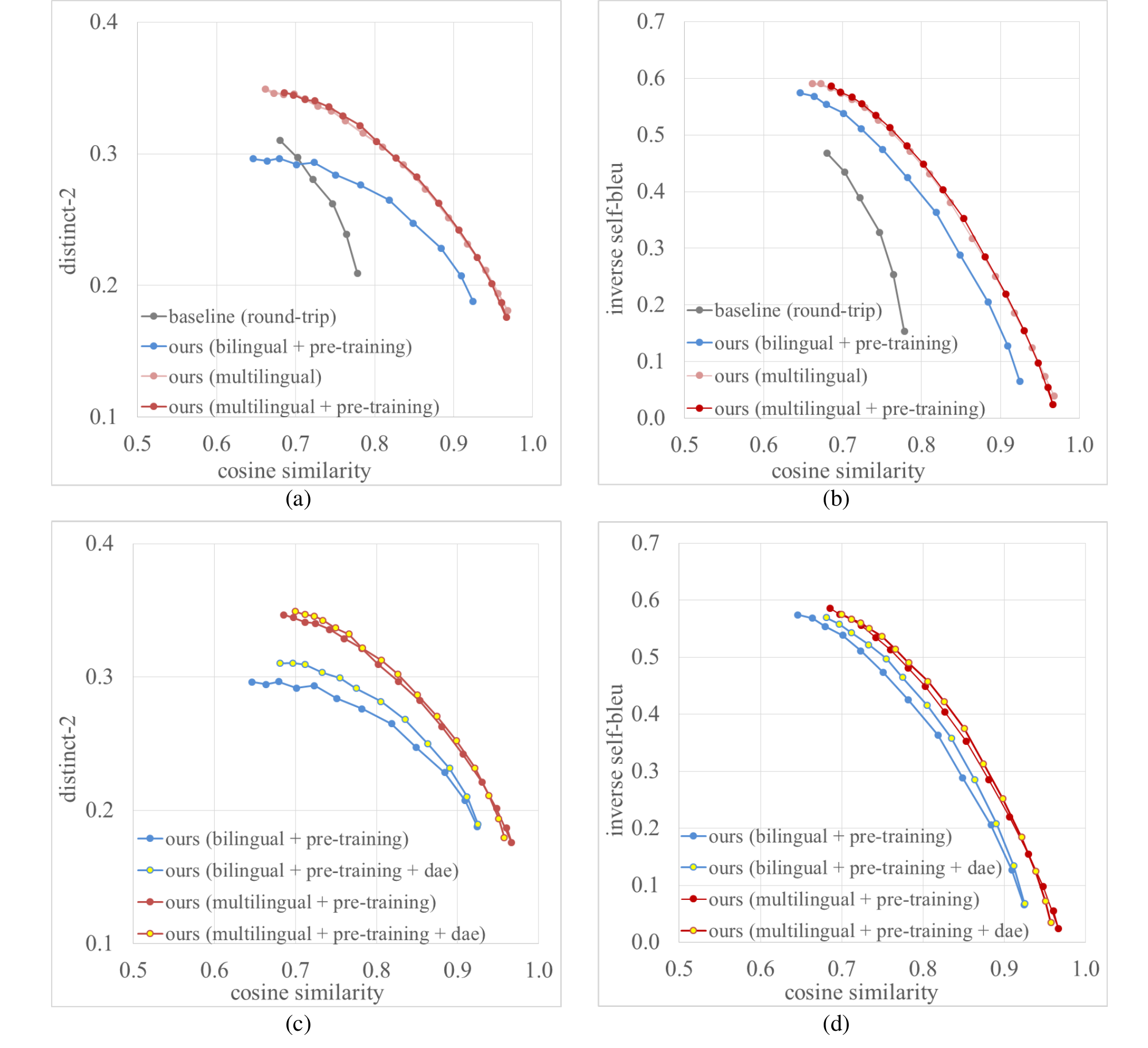} % Reduce the figure size so that it is slightly narrower than the column. Don't use precise values for figure width.This setup will avoid overfull boxes.
	\caption{Automatic evaluation: (a)(c) Distinct-2 versus Relevance; (b)(d) Inverse Self-BLEU versus Relevance.}
	\label{fig4}
\end{figure*}

\subsection{Datasets}

%Only text passages are kept for our experiments. We preprocess the corpus into sentence-level since we aim to tackle sentence-level paraphrase generation.
%On account of obtaining a multilingual paraphrasing model in general domain with written and spoken sentences,
We adopt the mixture of two multilingual translation corpus as our training data: MultiUN~\cite{Eisele2010} and OpenSubtitles~\cite{Lison2016}. MultiUN consists of 463,406 official documents in six languages, containing around 300M words for each language. OpenSubtitles is a corpus consisting of movie and TV subtitles, which contains 2.6B sentences over 60 languages. We select four shared languages of the two corpora: English, Spanish, Russian and Chinese. Statistics of the training corpus are shown in Table~\ref{table1}. Sentences are tokenized by Wordpiece as in BERT. A multilingual vocabulary of 50K tokens is used. For validation and testing, we randomly sample 10000 sentences respectively from each language pair. The rest data are used for training. For monolingual pre-training, we use English Wikipedia\footnote{\url{https://dumps.wikimedia.org/enwiki/latest/}} corpus, which contains 2,500M words.

\iffalse % Quora evaluation
For evaluation, except for test set from MultiUN and OpenSubtitles. We also adopts a labeled paraphrase corpus Quora Question Pairs\footnote{https://www.kaggle.com/c/quora-question-pairs} (QQP) to automatically evaluate quality of paraphrasing through calculating BLEU~\cite{Papineni2002} between generated sentence and its reference. QQP is a human annotated corpus indicating whether two sentences are paraphrase to each other. It offers 404K pairs of sentences with label.
\fi

\subsection{Experimental Settings}
We implement our model in Tensorflow~\cite{45381}. The size of our Transformer model is identical to BERT-base~\cite{Devlin2019}.
The model is constituted by 12 layers of Transformer blocks. Number of dimension of token embedding, position embedding and transformer hidden state are 768, while that of states in position-wise feed-forward networks are 3072. The number of attention heads is 12. Models are train using Adam optimization \cite{Kingma2015} with a learning rate up to \(1e-4\), \(\beta_1=0.9\), \(\beta_2=0.999\) and \(L2\) weight decay of 0.01. We use top-k truncated random sampling strategy for inference that only sample from k candidate words with highest probabilities.

Throughout our experiments, we train and evaluate two models for paraphrase generation: the bilingual model and the multilingual model. The bilingual models are trained only with English$\leftrightarrow$Chinese, while the multilingual models are trained with all the data between the four languages. The round-trip translation baseline is based on the Transformer-based neural translation model.
%We trained bilingual (English$\leftrightarrow$Chinese) and multilingual (all directions among English, Espanol, Russian and Chinese) in various settings. Then compare them with the baseline round-trip translation method which is based on our best translation model.

\subsection{Automatic Evaluation}
We evaluate the relevance between input and generated paraphrase as well as the diversity among multiple generated paraphrases from the same input. For relevance, we use the cosine similarity between the sentential representations~\cite{Liu2016}. Specifically, we use the Glove-840B embeddings \cite{Pennington2014} for word representation and Vector Extrema \cite{Liu2016} for sentential representation.
For generation diversity, we employ two evaluation metrics: Distinct-2\footnote{\url{https://github.com/neural-dialogue-metrics/Distinct-N}} and \textit{inverse} Self-BLEU (defined as: $1-$Self-BLEU)~\cite{Zhu2018}. Larger values of Distinct-2 and inverse Self-BLEU indicate higher diversity of the generation.
%In practice, we use \textit{inverse} Self-Bleu (inverse \(Self\-Bleu=1-Self\-Bleu\)) so that the higher its value is, the better diversity it represents. Evaluation results are revealed in Figure~\ref{fig4}.

%In Figure~\ref{fig4}, horizontal coordinates indicate \textbf{relevance} (cosine similarity), the best result lies on the rightmost. Vertical coordinates in Figure~\ref{fig4}-(a)/-(c) refer to \textbf{distinct-2}, while those in Figure~\ref{fig4}-(b)/-(d) refer to \textbf{Self-Bleu} diversity (\textit{inverse} Self-Bleu). The best results of diversity lie on the top. In sum, points lying on top right corner represent the best overall performance considering relevance and diversity simultaneously.

For each model, we draw curves in Figure~\ref{fig4} with the aforementioned metrics as coordinates, and each data-point is obtained at a specific sampling temperature. Since a good paraphrasing model should generate both relevant and diverse paraphrases, the model with curve lying towards the up-right corner is regarded as with good performance.

%In other words, we perform inference over test set using candidate models, in \textbf{a range of sampling temperatures}. For a model, Each temperature setting lead to a point in Figure~\ref{fig4}.

\subsubsection{Comparison with Baseline}
First we compare our models with the conventional pivoting method, i.e., round-trip translation. As shown in Figure~\ref{fig4} (a)(b), either the bilingual or the multilingual model is better than the baseline in terms of relevance and diversity in most cases. In other words, with the same generation diversity (measured by both Distinct-2 and Self-BLEU), our models can generate paraphrase with more semantically similarity to the input sentence.

Note that in Figure~\ref{fig4} (a), there is a cross point between the curve of the bilingual model and the baseline curve when relevance is around 0.71. We particularly investigate generated paraphrases around this point and find that the baseline actually achieves better relevance when Distinct-2 is at a high level ($>$0.3). It means our bilingual model is semantically drifting faster than the baseline model as the Distinct-2 diversity increases.
% The reason why this happens may be that the confidence of the baseline model's word distribution $P(y_j|z_1,...,z_h,y_1,...,y_{j-1})P(Z|X)$ (supervised translation model, $Z=[z_1,z_2,...,z_h]$ denotes a pivot sentence) is higher than our bilingual model's $P(y_j|x_1,...,x_n,y_1,...,y_{j-1})$ (zero-shot paraphrasing model).
The round-trip translation performs two-round of supervised translations, while the zero-shot paraphrasing performs single-round unsupervised `translation' (paraphrasing). We suspect that the unsupervised paraphrasing can be more sensitive to the decoding strategy. It also implies the latent, language-agnostic representation may be not well learned in our bilingual model. While on the other hand, our multilingual model alleviate this insufficiency. We further verify and analyze it as follows.

\subsubsection{Multilingual Models}
As mentioned above, our bilingual model can be unstable in some cases due to the lack of a well-learned language-agnostic semantic representation. A natural method is to introduce multilingual corpus, which consists of various translation directions. Training over multilingual corpus forces the model to decouple the language type and semantic representation.

Empirical results shows that our multilingual model performs significantly better than the bilingual model. The red and blue curves in Figure~\ref{fig4} (a)(b) demonstrates a great improvement of our multilingual model over the bilingual model. In addition, the multilingual model also significantly outperforms the baseline in the setting with the reasonable relevance scores.

% For instance, given an input \textit{"I got you, girl."}, we samples 3 sentences from each model which achieve identical distinct-2. The \textbf{baseline} generates [\textit{"I'll help you."}, \textit{"Coming, girl."}, \textit{"I gotcha."}], while \textbf{our bilingual} model generates [\textit{"I got you, girl. I got ya, baby. I got ya! Come on!"}, \textit{"-I got you. -Hey, girl."}, \textit{"I got ya, kid"}] and \textbf{our multilingual} model generates [\textit{"Gotcha, girl."}, \textit{"Got you girl."}, \textit{"I got you, girl."}]. It is obvious that our multilingual model performs the best considering relevance and diversity simultaneously.

%\begin{figure*}[htb]
%	\centering
%	\includegraphics[width=2\columnwidth]{figure5} % Reduce the figure size so that it is slightly narrower than the column. Don't use precise values for figure width.This setup will avoid overfull boxes.
%	\caption{Paraphrases generated by different models. For each input source, we randomly sample three paraphrases from the each model.}
%	\label{fig5}
%\end{figure*}

\subsubsection{Denoising Auto-Encoder}
To verify the effectiveness of DAE in our model, various experiments with different hyper-parameters were conducted. We find that DAE works the best when uniformly perturbing input sentences with probability 0.01, using only \textit{Deletion} and \textit{Reordering} operations. We investigate DAE over both bilingual and multilingual models as plotted in Figure~\ref{fig4} (c)(d). Curves with the yellow circles represent models with DAE training.

Results in the Figure~\ref{fig4} (c)(d) demonstrate positive effects of DAE in either bilingual or multilingual models. It is worth to note that, while DAE have marginal impact on multilingual model, it improves bilingual model significantly. This is an evidence indicating that DAE can improve the model in learning a more robust representation.

More specifically, since \textit{Deletion} forces model to focus on sentence-level semantics rather than word-level meaning while \textit{Reordering} forces model to focus more on meaning rather than their positions, it would be more difficult for a model to learn shortcuts (e.g. copy words). In other words, DAE improves models' capability in extracting deep semantic representation, which has a similar effect to introducing multilingual data.

\subsubsection{Monolingual Pre-Training}

\begin{table}[ht]
	\caption{Log-probabilities of the generated sentences. \(\surd\) and \(\times\) symbols denote learning with or without pre-training respectively, \textbf{bold} font denotes greater values.}\smallskip
	\centering
	\resizebox{1\columnwidth}{!}{
		\smallskip\begin{tabular}{l|c|c|c}
			\hline
			Model & Sampling & Pre-Training & Log-Prob  \\\hline
			\multirow{6}{*}{Multilingual} & \multirow{2}{*}{greedy, temp=1} & \multicolumn{1}{c|}{\(\surd\)} & \multicolumn{1}{c}{\textbf{-0.1427}} \\\cline{3-4}
			&  & \multicolumn{1}{c|}{\(\times\)} & \multicolumn{1}{c}{-0.1428} \\\cline{2-4}
			& \multirow{2}{*}{top-3, temp=1} & \multicolumn{1}{c|}{\(\surd\)} & \multicolumn{1}{c}{\textbf{-0.1425}} \\\cline{3-4}
			&  & \multicolumn{1}{c|}{\(\times\)} & \multicolumn{1}{c}{-0.1448} \\\cline{2-4}
			& \multirow{2}{*}{top-3, temp=1.5} & \multicolumn{1}{c|}{\(\surd\)} & \multicolumn{1}{c}{\textbf{-0.1420}} \\\cline{3-4}
			&  & \multicolumn{1}{c|}{\(\times\)} & \multicolumn{1}{c}{-0.1425} \\\hline
			\multirow{6}{*}{Bilingual} & \multirow{2}{*}{greedy, temp=1} & \multicolumn{1}{c|}{\(\surd\)} & \multicolumn{1}{c}{\textbf{-0.1472}} \\\cline{3-4}
			&  & \multicolumn{1}{c|}{\(\times\)} & \multicolumn{1}{c}{-0.1484} \\\cline{2-4}
			& \multirow{2}{*}{top-3, temp=1} & \multicolumn{1}{c|}{\(\surd\)} & \multicolumn{1}{c}{\textbf{-0.1487}} \\\cline{3-4}
			&  & \multicolumn{1}{c|}{\(\times\)} & \multicolumn{1}{c}{-0.1502} \\\cline{2-4}
			& \multirow{2}{*}{top-3, temp=1.5} & \multicolumn{1}{c|}{\(\surd\)} & \multicolumn{1}{c}{\textbf{-0.1461}} \\\cline{3-4}
			&  & \multicolumn{1}{c|}{\(\times\)} & \multicolumn{1}{c}{-0.1506} \\\hline
		\end{tabular}
	}
	\label{table2}
\end{table}
As shown in Figure~\ref{fig4} (a)(b), the model with language model pre-training almost performs equally to its contemporary without pre-training. However, evaluations on fluency uncover the value of pre-training. We evaluate a group of models over our test set in terms of fluency, using a n-grams language model\footnote{\url{http://www.openslr.org/11/}} trained on 14k public domain books.

As depicted in Table~\ref{table2}%\footnote{\(\circ\) and \(\times\) symbols denote training with or without pre-training respectively, \textbf{bold} font denotes greater values.}
, models with language model pre-training stably achieves greater log-probabilities than the model without pre-training. Namely, language model pre-training brings better fluency.

\iffalse
\subsubsection{Unsupervised Performance}
BLEU scores are calculated using Moses~\cite{Koehn:2007:MOS:1557769.1557821} multi-bleu.perl script.
\begin{table}[ht]
	\caption{BLEU scores on Quora dataset.}\smallskip
	\centering
	\resizebox{.8\columnwidth}{!}{
		\smallskip\begin{tabular}{l|c}
			\hline
			\textbf{Models} & \textbf{BLEU} \\
			\hline
			{DNPG~\cite{Li2019}} & 10.00  \\
			\hline
			{Adapted-DNPG~\cite{Li2019}} & 16.98  \\
			\hline
			{Round-trip} & 20.08  \\
			\hline
			\hline
			{Bilingual (ours)} & \textbf{20.29}  \\
			\hline
			{Multilingual (ours)} & 20.02  \\
			\hline
			{Bilingual + Pre-training (ours)} & 18.02  \\
			\hline
			{Multilingual + Pre-training (ours)} & 19.36  \\
			\hline
		\end{tabular}
	}
	\label{table3}
\end{table}
\fi

\subsection{Human Evaluation}
200 sentences are sampled from our test set for human evaluation. The human evaluation guidance generally follows that of \cite{Li2018} but with a compressed scoring range from [1, 5] to [1, 4]. We recruit five human annotators to evaluate models in semantic relevance and fluency. A test example consists of one input sentence, one generated sentence from baseline model and one generated sentence from our model. We randomly permute a pair of generated sentences to reduce annotators' bias on a certain model. Each example is evaluated by two annotators.

\begin{table}[ht]
	\caption{Human evaluation results.}
	\centering
	\resizebox{1\columnwidth}{!}{
		\begin{tabular}{lccc}
			\hline
			Model & Relevance & Fluency & Agreement \\
			\hline
			Round-trip & 2.72 & 3.61 & 0.36 \\
			Multilingual (ours) & \textbf{3.43} & \textbf{3.75} & 0.35 \\
			\hline
		\end{tabular}
	}
	\label{table4}
\end{table}

As shown in Table~\ref{table4}, our method outperforms the baseline in both relevance and fluency significantly. We further calculate agreement (Cohen's kappa) between two annotators.

Both round-trip translation and our method performs well as to fluency. But the huge gap of relevance between the two systems draw much attention of us. We investigate the test set in details and find that round-trip approach indeed generate more noise as shown in case studies.

\subsection{Case Studies}
\begin{table*}[htbp]
	\small
	\centering
	\caption{Case studies. For each input source, we randomly sample three paraphrases for comparison.}\smallskip
	\begin{tabular}{l|l}
		\hline
		Source                                                                         & I guess I kinda felt insignificant.                                                      \\ \hline
		\multirow{3}{*}{Round-trip}                                                    & \textit{I think I just don't feel right about that.}                                                     \\ %\cline{2-2}
		& \textit{I guess I'm a little uncomfortable.}                                                             \\ %\cline{2-2}
		& \textit{I think I'm a little bit of a problem right now.}                                                \\ \hline
		\multirow{3}{*}{\begin{tabular}[c]{@{}l@{}}Multilingual\\ (ours)\end{tabular}} & \textit{I guess I was feeling a bit unsignificant.}                                                      \\ %\cline{2-2}
		& \textit{I guess I felt some sorts of insignificant. }                                                    \\ %\cline{2-2}
		& \textit{I guess I kind of felt insignificant. }                                                         \\ \hline
		Source                                                                         & This site will make better use of the guide and will increase its distribution.                 \\ \hline
		\multirow{3}{*}{Round-trip}                                                    & \textit{The site would make better use of the guidelines and would be expanded.}                         \\ %\cline{2-2}
		& \textit{The site will make the best use of guides and expand them. }                                     \\ %\cline{2-2}
		& \textit{This site would have made use more of the guidelines and could be expanded to its distribution.} \\ \hline
		\multirow{3}{*}{\begin{tabular}[c]{@{}l@{}}Multilingual\\ (ours)\end{tabular}} & \textit{This web site will make better use of the guide and will increase its dissemination.}            \\ %\cline{2-2}
		& \textit{This site will better utilize the guide, and will improve its distribution.}                     \\ %\cline{2-2}
		& \textit{The web site is going to make the guide's use more efficient and its distribution will grow.}    \\ \hline
		Source                                                                         & That's how eric got the passcodes.                                                              \\ \hline
		\multirow{3}{*}{Round-trip}                                                    & \textit{Then eric has a code.}                                                                           \\ %\cline{2-2}
		& \textit{Then eric has the codes.}                                                                        \\ %\cline{2-2}
		& \textit{Then erik'll have the codes.}                                                                    \\ \hline
		\multirow{3}{*}{\begin{tabular}[c]{@{}l@{}}Multilingual\\ (ours)\end{tabular}} & \textit{That's the way eric got the password codes.}                                                     \\ %\cline{2-2}
		& \textit{That's how eric got passwords.}                                                                  \\ %\cline{2-2}
		& \textit{That's where eric gets the passcodes.}                                                           \\ \hline
	\end{tabular}
	\label{table5}
\end{table*}

We further study some generated cases from different models. All results in Table~\ref{table5} are generated over our test set using randomly sampling. For both baseline and multilingual model, we tune their sampling temperatures to control the Distinct-2 and the inverse Self-BLEU at 0.31 and 0.47 respectively.

In the case studies, we find that our method usually generates sentences with better relevance to source inputs, while the round-trip translation method can sometimes run into serious semantic drift. In the second case, our model demonstrates a good feature that it maintains the meaning and even a proper noun \(guide\) unchanged while modifies the source sentence by both changing and reordering words. This feature may be introduced by DAE perturbation strategies which improves model's robustness and diversity simultaneously. These results evidence that our methods outperforms the baseline in both relevance and diversity.

\section{Related Work}
Generating paraphrases based on deep neural networks, especially Seq2Seq models, has become the mainstream approach. A majority of neural paraphrasing models tried to improve generation quality and diversity with high-quality paraphrase corpora. \citet{prakash-etal-2016-neural} starts a deep learning line of paraphrase generation through introducing stacked residual LSTM network. A word constraint model proposed by \citet{Cao2017} improves both generation quality and diversity. \citet{Gupta2018} adopts variational auto-encoder to further improve generation diversity. \citet{Li2018} utilize neural reinforcement learning and adversarial training to promote generation quality. \citet{Li2019} decompose paraphrase generation into phrase-level and sentence-level.

Several works tried to generate paraphrases from monolingual non-parallel or translation corpora.  \citet{Zhang2016} exploits Markov Network model to extract paraphrase tables from monolingual corpus. \citet{quirk-etal-2004-monolingual}, \citet{Wubben2010} and \citet{Wubben2014} create paraphrase corpus through clustering and aligning paraphrases from crawled articles or headlines. With parallel translation corpora, pivoting approaches such round-trip translation \cite{Mallinson2018} and back-translation \cite{Wieting2018} are explored.

However, to the best knowledge of us, none of these paraphrase generation models has been trained directly from parallel translation corpora as a single-round end-to-end model.

\section{Conclusions}
In this work, we have proposed a Transformer-based model for zero-shot paraphrase generation, which can leverage huge amount of off-the-shelf translation corpora. Moreover, we improve generation fluency of our model with language model pre-training. Empirical results from both automatic and human evaluation demonstrate that our model surpasses the conventional pivoting approaches in terms of relevance, diversity, fluency and efficiency. Nevertheless, there are some interesting directions to be explored. For instance, how to obtain a better latent semantic representation with multi-modal data and how to further improve the generation diversity without sacrificing relevance. We plan to strike these challenging yet valuable problems in the future.
% Moreover, we even outperforms a state-of-the-art model trained on paraphrase corpus under fully unsupervised setting.

% \section{Acknowledgements}

\bibliography{con_papers,jour_papers,arx_papers,books}
\bibliographystyle{acl_natbib}

\end{CJK*}
\end{document}